\documentclass[11pt]{article}
\usepackage[T1]{fontenc}
\usepackage[letterpaper,margin=0.85in,headheight=14pt]{geometry}
\usepackage{times,amsmath,amssymb,graphicx,booktabs,array,microtype}
\usepackage[dvipsnames]{xcolor}
\usepackage{tikz}
\usetikzlibrary{arrows.meta,positioning,fit}
\usepackage{enumitem,caption,fancyhdr,xurl,placeins,titlesec}
\usepackage[numbers,sort&compress]{natbib}
\usepackage[colorlinks=true,linkcolor=MidnightBlue,citecolor=MidnightBlue,urlcolor=MidnightBlue]{hyperref}
\setlist{itemsep=2pt,topsep=4pt,leftmargin=1.6em}
\titlespacing*{\section}{0pt}{14pt plus 2pt minus 2pt}{6pt plus 1pt minus 1pt}
\titlespacing*{\subsection}{0pt}{10pt plus 2pt minus 2pt}{4pt plus 1pt minus 1pt}
\titlespacing*{\paragraph}{0pt}{6pt plus 1pt minus 1pt}{1em}
\newcommand{\rms}{\operatorname{RMS}}
\newcommand{\attn}{\operatorname{Attn}}
\newcommand{\bref}{B_{\mathrm{ref}}}
\newcommand{\code}[1]{\nolinkurl{#1}}
\newcommand{\smallnote}[1]{\par\vspace{2pt}{\footnotesize #1\par}}
\makeatletter
\renewcommand{\@maketitle}{%
  \begin{center}
    {\LARGE\@title\par}\vspace{12pt}
    {\large\@author\par}\vspace{4pt}
  \end{center}\vspace{6pt}}
\makeatother
\title{\textbf{KITE: KV-Invariant Transformer Expansion\\for Efficient Agentic LLM Scaling}}
\author{%
  \normalsize
  \begin{tabular}{@{}c@{}}
    Zhiheng Hu\quad Yixun Wei\quad Jian Zhou\quad Yizhuang Zhou\quad Ji Li\quad Xing Chen\\[2pt]
    Yang Li\quad Bojun Wang\quad Yibo Zhu\quad Xiangyu Zhang\quad Daxin Jiang\\[4pt]
    StepFun\hspace{0.3em}\raisebox{-0.25em}{\includegraphics[height=1.2em]{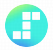}}
  \end{tabular}%
}

\begin{document}
\maketitle
\thispagestyle{empty}
\begin{abstract}
Scaling a language model is not only a question of final quality: the architectural choice determines how much computation is spent during training, prompt processing, and autoregressive decoding to achieve certain model quality.
An ideal model architecture should lower all above computation costs to facilitate scaling to a larger model, while ensure the larger model indeed outperforms smaller baselines.
We introduce \emph{KV-Invariant Transformer Expansion (KITE)}, a scaling paradigm that achieves this goal. It trains the model from a smaller size to a larger size ({\em i.e.,} saving training costs via upcycling), while places newly added parameters in regions that do not affect attention KV. Consequently, during inference, prefilling KV only relies on the smaller part of the model, so the inference costs are saved. As a concrete instantiation, we present Step Scale Transformer (SST), a two-tower decoder in which one tower produces KV and the other reads them. 
At comparable cumulative training compute, SST, a 67B MoE model with 2.15B active body parameters per decode token, achieves lower training loss than 47B and 63B MoE Transformers with 1.48B and 2.02B active body parameters, respectively, while reducing estimated inference cost by 6.7\% and 31.6\%.
\end{abstract}

\begin{figure}[!ht]
\centering\includegraphics[width=\linewidth]{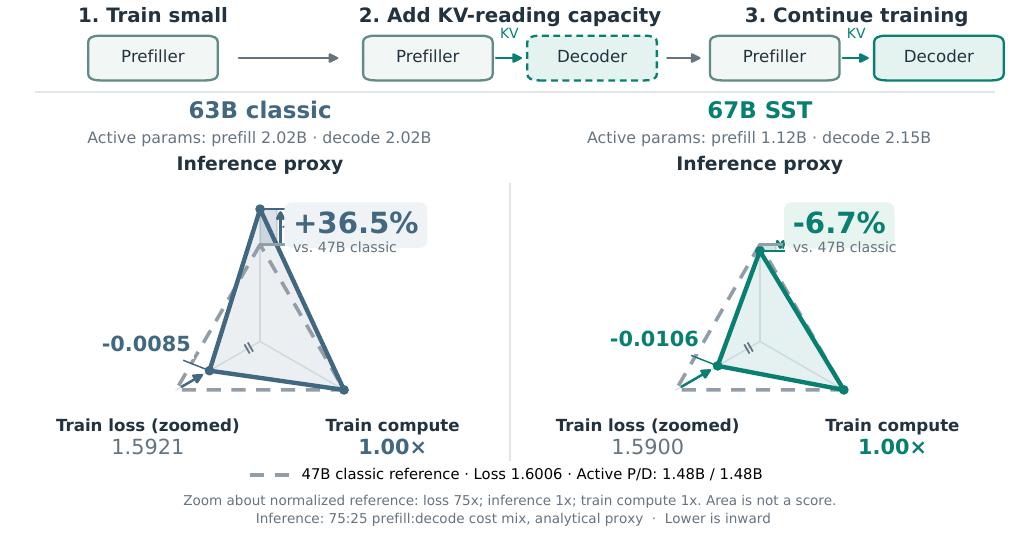}
\caption{KITE: training route and observed trade-offs. Both panels use the same 47B transformer trained from scratch as the reference; lower is inward. Active counts exclude embedding/head. The inference proxy is $0.75p+0.25d$, where $p$ and $d$ are bulk-prefill and decode active-body counts normalized to the reference.}
\label{fig:scaling-route}
\end{figure}
\clearpage

\section{Introduction}
\label{sec:intro}

Large language models (LLMs) exhibit strong scaling behavior: increasing model scale continues to bring substantial gains in model capability~\cite{kaplan2020scaling,hoffmann2022training}. This property is both one of the most important blessings of LLMs and a curse, as larger models require enormous compute budgets for both training and inference.

Consequently, one of the central problems in LLM research is how to scale models at lower compute cost. This goal is not only about reducing expenditure; it enables a larger model to be trained or served under the same compute budget, yielding a stronger model. MoE feed-forward layers provide a successful example: replacing a dense FFN with sparse expert routing substantially reduces training and inference computation at a fixed parameter count, or, equivalently, allows model capacity to grow at a fixed compute budget~\cite{shazeer2017moe,fedus2021switch}.

What is easy to miss in this success is that MoE reduces training and inference computation simultaneously. A saving on only one side may not improve the overall trade-off. For example, progressive upcycling trains a small model first and grows it into the target model partway through training, so much of the token budget runs on a cheaper graph~\cite{chen2015net2net,yu2026sparkling}. However, upcycled models are generally perceived to underperform same-size models trained
from scratch, so the training saving is paid back at inference,
where one serves the full target model yet obtains less than
the full target quality. KV reuse pays
on the other side. By letting deeper layers reuse the keys and values computed by shallower
layers, YOCO family keeps prefill affordable as a model grows \cite{sun2024yoco,wechat2026welm,dsv41}; Yet removing KV-producing
layers can sacrifice model quality; recovering the same quality may require a larger model or a longer training, which is effectively a training-cost loss. In both cases the gain on one axis is offset by a regression on the other. We therefore treat
the two costs jointly: we pursue a scaling paradigm whose model quality, training cost, and inference cost all
move in its favor.

This paper proposes \emph{KV-Invariant Transformer Expansion (KITE)}, a model-scaling paradigm. Its core idea is that when the model is scaled, it is divided into two regions: the region that determines the key--value (KV) tensors, and the region that does not affect them. Scaling is carried out by
expanding capacity in the region that does not affect the KV, which yields a good ratio of model
intelligence to both training and inference cost. That is because the added capacity
appears partway through training rather than from token zero, so the quality it buys comes
at a favorable training cost. Meanwhile, this scaling does not affect the KV computation, so
the prefill cost at inference remains unchanged. 

KITE leverages the fact that prompt processing is significant in agentic workloads, where repeated model calls process accumulated context and new tool outputs~\cite{yao2022react,yang2024sweagent}. Even with prompt caching, substantial uncached input remains. Public OpenRouter traffic illustrates this input-heavy demand: across the six models in Table~\ref{tab:openrouter-intro}, uncached input exceeds output in token volume and accounts for most estimated uncached-input and output charges. Consequently, by saving on the prefill costs, KITE, like YOCO, produces models that are overall much cheaper to serve than classic transformers.

\begin{table}[!htb]
\centering
\small
\setlength{\tabcolsep}{5pt}
\caption{Input demand in selected OpenRouter traffic.
Input includes cached tokens; output includes reasoning tokens.
Charge share is uncached-input charges divided by uncached-input
plus output charges, excluding cache-read charges.
Estimates aggregate available standard and Batch activity within
August 22--September 20, 2026 (UTC), using September 21 base prices.
Details appear in Appendix~\ref{app:openrouter}.}
\label{tab:openrouter-intro}
\begin{tabular}{@{}lrrr@{}}
\toprule
Model &
\shortstack{Input/output\\tokens} &
\shortstack{Uncached input/\\output tokens} &
\shortstack{Uncached-input\\charge share} \\
\midrule
GPT-6 Astra      & $106.43\times$ & $14.64\times$ & 74.6\% \\
GPT-6 Astra Pro  & $63.15\times$  & $14.01\times$ & 73.8\% \\
Claude Fable 5   & $63.38\times$  & $16.29\times$ & 76.7\% \\
Claude Sonnet 5 & $59.88\times$ & $12.12\times$ & 70.8\% \\
Gemini 3.7 Flash & $37.79\times$  & $10.15\times$ & 67.4\% \\
Gemini 3.8 Flash & $46.14\times$  & $10.51\times$ & 67.9\% \\
\bottomrule
\end{tabular}
\end{table}

To instantiate KITE, we design a straightforward architecture: \emph{Step Scale Transformer} (SST). SST contains two same-depth towers that execute sequentially. The Prefiller produces layer-wise KV, which the Decoder reuses to predict the next token. During generation, bulk prefill uses only the first tower, while the prompt-boundary prediction and decode use both. The source model is trained first; after expansion, both towers continue training jointly. Section~\ref{sec:architecture} provides the architecture and execution details.

We spend the saved compute on scaling the model up moderately, just as MoE does. Compared
with a Transformer baseline trained from scratch, this yields a net gain in the three-way
trade-off among training cost, inference cost, and model intelligence.
Figure~\ref{fig:scaling-route} illustrates these trade-offs. KITE can also be used together with
earlier successful designs such as MoE and hybrid attention; they are orthogonal and do not
interfere with one another. Although SST is probably not the best KITE
design, it is straightforward and suffices to demonstrate the positive effect of the
paradigm. We are continuing to explore this paradigm, and we anticipate that more and more
related designs can emerge.

To make the comparison clear, we build a scaling ladder across multiple model scaling paradigms. The tokenizer, data recipe, training protocol, and evaluation sets are kept identical, and we trace the cost--quality relationship of each scaling paradigm. A better scaling paradigm is
one that reaches a lower training loss under a comparable training-FLOPs budget while also incurring a lower inference cost. The experiments show that the SST instantiation of KITE, expanded to 67B parameters, reaches an EMA-200 training loss of 1.5900, 0.0106 lower than 47B classic (1.6006) and 0.0021 lower than 63B classic (1.5921). At the final training endpoints, SST and 63B classic both use approximately 100\% of the 47B baseline's theoretical cumulative training FLOPs, including both SST training stages. With illustrative cost weights of 75\% for prefill and 25\% for decode, SST's analytical inference-cost proxy is 6.7\% lower than 47B classic and 31.6\% lower than 63B classic (Figure~\ref{fig:scaling-route}). On downstream benchmarks, SST scores higher than both baselines on OpenBookQA, MMLU, GSM8K, MATH, HumanEval, MBPP, and BBH (Table~\ref{tab:bmk}). 

\section{KITE: KV-Invariant Transformer Expansion}
\label{sec:architecture}

\begin{figure}[!htb]
\centering\resizebox{0.94\linewidth}{!}{\begin{tikzpicture}[x=1cm,y=1cm,font=\sffamily\small,>=Stealth,
 box/.style={rounded corners=2pt,minimum width=3.1cm,minimum height=.62cm,align=center,line width=.7pt},
 h2style/.style={box,draw=teal!70!black,fill=teal!7},
 h1style/.style={box,draw=blue!65!black,fill=blue!5},
 neutral/.style={box,draw=black!35,fill=black!2},
 flow/.style={->,line width=.9pt,draw=black!60},
 kv/.style={->,line width=1pt,draw=teal!75!black,preaction={draw=white,line width=3pt}},
 state/.style={->,rounded corners=4pt,line width=.9pt,draw=orange!75!black},
 note/.style={font=\sffamily\footnotesize,text=black!65,align=center}]

\draw[rounded corners=3pt,draw=black!18,line width=.6pt] (-8.25,-1.75) rectangle (-3.65,8.02);
\draw[rounded corners=3pt,draw=black!18,line width=.6pt] (-.22,-1.75) rectangle (10.60,8.02);
\draw[rounded corners=3pt,draw=teal!30!black,line width=.5pt] (-.02,.48) rectangle (3.62,5.62);
\draw[rounded corners=3pt,draw=blue!65!black,dashed,line width=.7pt] (6.38,-.66) rectangle (10.02,5.62);
\node[font=\sffamily\bfseries\large] at (-6,9.04) {Stage 1};
\node[font=\sffamily\small,text=teal!65!black] at (-6,8.55) {Train the Prefiller};
\node[font=\sffamily\bfseries\large] at (5.19,9.04) {Stage 2};
\node[font=\sffamily\small,text=teal!65!black] at (5.19,8.55) {Jointly train both towers};

\node[font=\sffamily\bfseries,text=teal!65!black] at (-6,5.25) {Prefiller};
\node[h2style] (s1) at (-6,1.0) {Transformer block 1};
\node[h2style] (s2) at (-6,2.15) {Transformer block 2};
\node[h2style] (sL) at (-6,4.1) {Transformer block $L$};
\node at (-6,3.12) {$\vdots$};
\node[neutral] (semb) at (-6,-1.2) {Token embedding};
\node[neutral,minimum width=3.45cm] (shead) at (-6,7.02) {Output norm + LM head};
\node[note] at (-6,7.72) {Next-token prediction};
\draw[flow] (semb)--(s1);
\draw[flow] (s1)--(s2);
\draw[flow] (s2.north)--++(0,.32);
\draw[flow] (sL.south)++(0,-.35)--(sL.south);
\draw[flow] (sL.north)--(-6,4.67)--(-3.83,4.67)--(-3.83,7.02)--(shead.east);
\draw[flow] (shead.north)--++(0,.27);
\node[font=\sffamily\bfseries] at (-2.12,3.55) {Expand};
\draw[->,line width=1.3pt,draw=black!65] (-3.40,2.95)--(-.60,2.95);
\node[note] at (-2.12,2.20) {Initialize Decoder\\from source weights};
\node[note,font=\sffamily\scriptsize] at (1.8,4.85) {retained};
\node[note,font=\sffamily\scriptsize] at (8.2,4.85) {added at expansion};

\node[font=\sffamily\bfseries,text=teal!65!black] at (1.8,5.25) {Prefiller};
\node[font=\sffamily\bfseries,text=blue!65!black] at (8.2,5.25) {Decoder};
\node[h2style] (p1) at (1.8,1.0) {Transformer block 1};
\node[h2style] (p2) at (1.8,2.15) {Transformer block 2};
\node[h2style] (pL) at (1.8,4.1) {Transformer block $L$};
\node[h1style] (c1) at (8.2,1.0) {Transformer block 1};
\node[h1style] (c2) at (8.2,2.15) {Transformer block 2};
\node[h1style] (cL) at (8.2,4.1) {Transformer block $L$};
\node at (1.8,3.12) {$\vdots$};
\node at (8.2,3.12) {$\vdots$};
\node[neutral] (emb) at (1.8,-1.2) {Token embedding};
\node[neutral,draw=orange!65!black,minimum height=.68cm] (bridge) at (8.2,-.15) {Entry bridge};
\node[neutral,minimum width=3.45cm] (readout) at (5,5.95) {RMSNorm (no affine)};
\node[neutral,minimum width=3.45cm] (head) at (5,7.02) {Output norm + LM head};
\node[note] at (5,7.72) {Next-token prediction};

\draw[flow] (emb)--(p1);
\draw[flow] (p1)--(p2);
\draw[flow] (p2.north)--++(0,.32);
\draw[flow] (pL.south)++(0,-.35)--(pL.south);
\draw[flow] (bridge)--(c1);
\draw[flow] (c1)--(c2);
\draw[flow] (c2.north)--++(0,.32);
\draw[flow] (cL.south)++(0,-.35)--(cL.south);

\coordinate (pfinal) at (1.8,4.67);
\draw[flow,-] (pL.north)--(pfinal);
\draw[flow] (cL.north)--(8.2,4.67)--(10.25,4.67)--(10.25,5.95)--(readout.east);
\draw[flow] (readout)--(head);
\draw[flow] (head.north)--++(0,.27);
\draw[state] (pfinal)--(4.55,4.67)--(4.55,-.15)--(bridge.west);
\fill[orange!75!black] (pfinal) circle (1.5pt);
\node[note,text=orange!75!black,fill=white,inner sep=2pt] at (4.65,4.98) {Final Prefiller state};
\draw[state] (emb.east)--(8.2,-1.2)--(bridge.south);
\node[note,text=orange!75!black,fill=white,inner sep=1pt] at (5,-1.2) {Embedding};

\draw[kv] (p1.east)--(c1.west) node[pos=.62,above,fill=white,inner sep=1pt,font=\footnotesize] {$K_1,V_1$};
\draw[kv] (p2.east)--(c2.west) node[pos=.62,above,fill=white,inner sep=1pt,font=\footnotesize] {$K_2,V_2$};
\draw[kv] (pL.east)--(cL.west) node[pos=.62,above,fill=white,inner sep=1pt,font=\footnotesize] {$K_L,V_L$};
\node[note,text=teal!65!black,fill=white,inner sep=2pt] at (5.55,3.05) {Layer-wise\\KV reuse};

\draw[->,draw=teal!75!black,line width=1pt] (.2,-2.3)--(.9,-2.3);
\node[note,anchor=west] at (1.0,-2.3) {Shared KV};
\draw[->,draw=orange!75!black,line width=.9pt] (3.3,-2.3)--(4.0,-2.3);
\node[note,anchor=west] at (4.1,-2.3) {Token-local bridge};
\draw[blue!65!black,dashed,rounded corners=1pt,line width=.7pt] (7.5,-2.47) rectangle (7.9,-2.13);
\node[note,anchor=west] at (8.05,-2.3) {New Decoder};
\end{tikzpicture}}
\caption{KITE instantiated with SST. Stage 1 trains the Prefiller; expansion adds a Decoder initialized from the source weights, and Stage 2 continues training both towers jointly. Within Stage 2, the Prefiller executes in full before the Decoder. Each Decoder block uses aligned Prefiller KV. A token-local entry bridge combines the separately normalized embedding and final Prefiller state. The final Decoder state is normalized and passed to the shared output norm and LM head for prediction.}
\label{fig:architecture}
\end{figure}
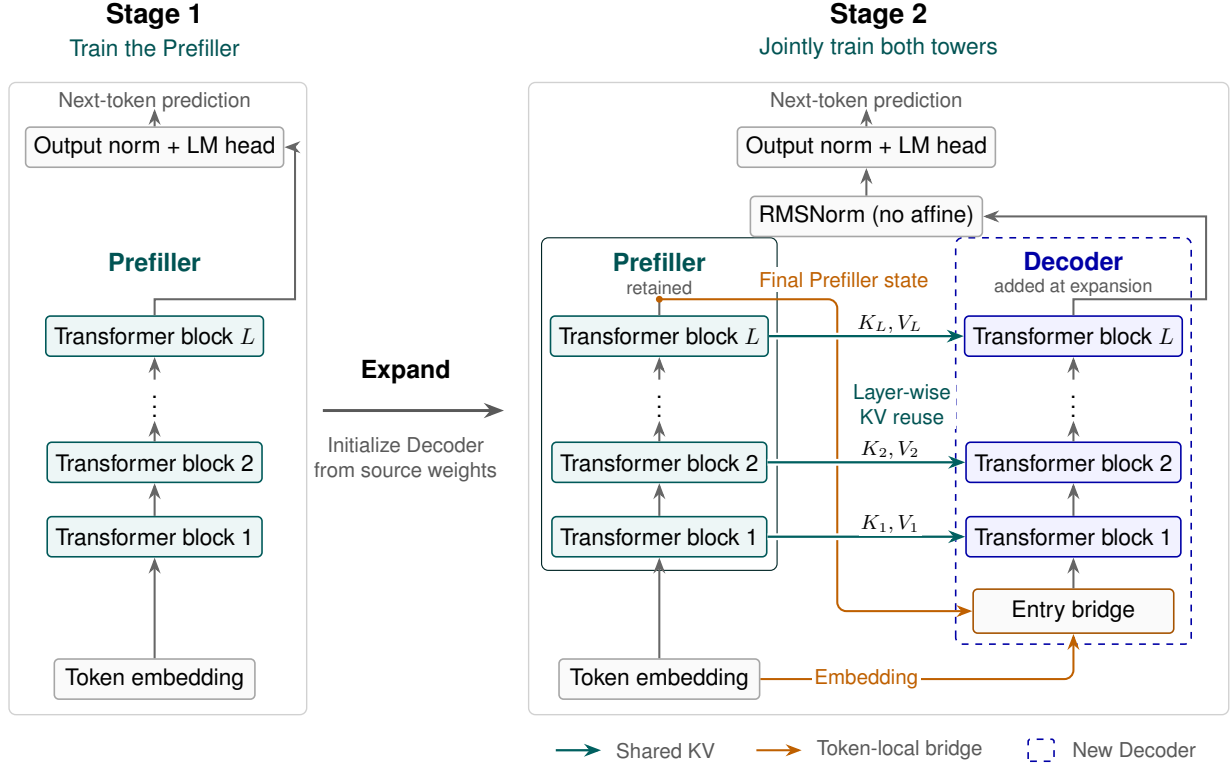

We illustrate KITE through SST, a concrete instantiation that adds KV-reading capacity to a smaller model and continues training both parts jointly (Figure~\ref{fig:architecture}).

\subsection{The design principle}

Let a causal model have a Prefiller $P$ and a Decoder $D$:
\begin{equation}
 (M_t,z_t)=P(x_{1:t};\theta_P),\qquad
 \ell_t=D(x_t,z_t,M_t;\theta_D),
\label{eq:factorization}
\end{equation}
where $M_t$ is reusable attention memory, $z_t$ denotes optional Prefiller features at position $t$, and $\ell_t$ gives next-token logits. After an initial small-model training stage, expansion adds or enlarges $D$ using the intermediate checkpoint, retaining the Prefiller graph and memory interface. Both parts then continue training.

The invariant is structural: neither weights nor KV values are frozen. The Decoder must not produce memory needed by future positions. To omit historical Decoder computation during generation, its retained positions must also be independent of omitted Decoder positions given the Prefiller outputs. These restrictions let prediction capacity grow without increasing \emph{bulk prefill}, the prompt-wide KV construction performed by $P$. The final-position readout for the first output token and the Decoder's work during decode remain necessary.

\subsection{Step Scale Transformer}

SST instantiates this principle with two Transformer towers~\cite{vaswani2017attention}: the Prefiller (superscript $P$) and the Decoder ($D$), each with 18 layers in our experiments. The Prefiller executes first and produces layer-wise KV. Each Decoder block computes its own query and reads KV from the corresponding Prefiller layer:

\begin{equation}
 q^{D}_{l,t}=Q_l^{D}(h^{D}_{l-1,t}),\qquad
 a^{D}_{l,t}=\attn\!\left(q^{D}_{l,t},K_l^{P}[\mathcal I_l(t)],V_l^{P}[\mathcal I_l(t)]\right).
\label{eq:attention}
\end{equation}

Here, $Q_l^{D}$ includes query projection and normalization, and $\mathcal I_l(t)$ specifies the positions allowed by the original causal full/sliding attention mask. Both towers retain the source model's position IDs. The Decoder has independent block parameters; its residual and FFN/MoE operations are token-local.

In the expanded SST model shown on the right of Figure~\ref{fig:architecture}, a token-local entry bridge initializes the Decoder from the embedding and final Prefiller state. Prediction uses the Decoder's final state. Gradients flow to the Prefiller through both the entry bridge and the reused KV. Given the Prefiller outputs, each Decoder position is independent of earlier Decoder states.

\subsection{Training and inference}
\label{sec:execution}

\paragraph{Training.} After expansion, SST jointly trains both towers. The Prefiller processes the full sequence, and the Decoder computes every supervised position; gradients pass through the reused KV into the Prefiller. The initial small-model training stage precedes the addition of the Decoder.

\paragraph{Prefill.} The Prefiller processes the complete prompt $x_{1:n}$ and produces the layer-wise KV cache. The Decoder needs to compute only the final prompt position $n$, attending to the Prefiller's KV, to predict the first output token. Earlier Decoder positions can be omitted because its other operations are token-local.

\paragraph{Decode.} Each new token passes through both towers. The Prefiller extends the KV cache, and the Decoder reads this KV to form the next-token prediction.

\subsection{Decoder input and output}

For the shared embedding/stem $e_t$ and final Prefiller state $h^{P}_{L,t}$, SST uses the entry bridge
\begin{equation}
 h^{D}_{0,t}=\rms_{\epsilon}(e_t)+\rms_{\epsilon}(h^{P}_{L,t}),\qquad
 \rms_{\epsilon}(v)=\frac{v}{\sqrt{\operatorname{mean}(v^2)+\epsilon}}.
\label{eq:bridge}
\end{equation}
The Decoder's final state produces the next-token logits:
\begin{equation}
 \ell_t=W_{\mathrm{out}}\,\mathrm{Norm}_{\mathrm{out}}(h^{D}_{L,t}).
\label{eq:tail}
\end{equation}
Here, $\mathrm{Norm}_{\mathrm{out}}$ denotes the complete output normalization; its implementation is specified in Appendix~\ref{app:schedule}. The entry-bridge RMS operations have no trainable parameters and use $\epsilon=10^{-5}$. The towers share one embedding, output normalization, and output head; embedding and head weights are untied.

\section{Model construction and training}
\label{sec:training}

\subsection{From the source to the expanded model}

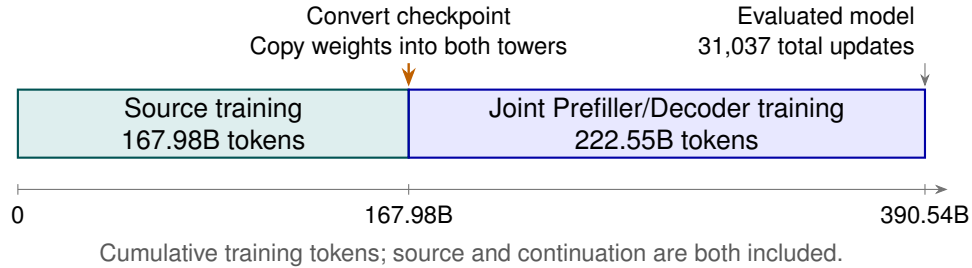
\begin{figure}[!htb]
\centering\begin{tikzpicture}[x=1cm,y=1cm,font=\sffamily\small,>=Stealth]
\fill[teal!13,rounded corners=2pt] (0,0) rectangle (5.17,.9);
\fill[blue!9,rounded corners=2pt] (5.17,0) rectangle (12,.9);
\draw[teal!65!black,line width=.8pt] (0,0) rectangle (5.17,.9);
\draw[blue!65!black,line width=.8pt] (5.17,0) rectangle (12,.9);
\node[align=center] at (2.585,.45) {Source training\\167.98B tokens};
\node[align=center] at (8.585,.45) {Joint Prefiller/Decoder training\\222.55B tokens};
\draw[->,black!55] (0,-.42)--(12.3,-.42);
\foreach \x/\lab in {0/0,5.17/167.98B,12/390.54B}{
  \draw[black!55] (\x,-.35)--(\x,-.49);
  \node[below,font=\sffamily\footnotesize] at (\x,-.5) {\lab};
}
\node[align=center,font=\sffamily\footnotesize] at (5.17,1.65) {Convert checkpoint\\Copy weights into both towers};
\draw[->,orange!75!black,line width=.9pt] (5.17,1.25)--(5.17,.94);
\node[anchor=east,align=right,font=\sffamily\footnotesize] at (12,1.65) {Evaluated model\\31,037 total updates};
\draw[->,black!55] (12,1.25)--(12,.94);
\node[below,font=\sffamily\footnotesize,text=black!65] at (6,-1) {Cumulative training tokens; source and continuation are both included.};
\end{tikzpicture}
\caption{Training route to the evaluated SST checkpoint: 13,350 source steps followed by 17,687 continuation steps.}
\label{fig:training-route}
\end{figure}

SST expands a 33.819B-parameter source into a 66.959B model while retaining the source tower as the Prefiller (Table~\ref{tab:config}). The evaluated route contains 167.98B source-stage tokens and 222.55B continuation tokens (Figure~\ref{fig:training-route}). Both stages contribute to the reported training budget.

The 47B and 63B classic-trained baselines are the next two scales above the 33.8B source in the same model family's scaling ladder, obtained by increasing depth and width while retaining the MoE design and attention pattern. The 47B model provides an intermediate-scale reference, and the 63B model provides a reference close to SST's 67B parameter count. 

\begin{table}[!htb]
\centering\small
\caption{Model configurations.}
\label{tab:config}
\begin{tabular}{@{}lrrrr@{}}
\toprule
Setting & Source & SST & 47B classic & 63B classic\\
\midrule
Layers & 18 & $18+18$ & 20 & 22\\
Hidden width & 2304 & 2304 & 2560 & 2816\\
MoE layers & 16 & $16+16$ & 18 & 20\\
Routed / selected experts & 512 / 8 & 512 / 8 & 512 / 8 & 512 / 8\\
Attention pattern & SSSF & SSSF & SSSF & SSSF\\
Total parameters (B) & 33.819 & 66.959 & 46.727 & 62.691\\
Decode active body parameters (B/token) & 1.120 & 2.155 & 1.477 & 2.016\\
\bottomrule
\end{tabular}
\smallnote{Each tower starts with two dense FFN layers. SSSF is three sliding layers followed by one full-attention layer, with an SF remainder for 18/22 layers. Counts describe the effective architecture; Appendix~\ref{app:params} gives the geometry and counting convention.}
\end{table}

\subsection{Expansion and continued training}

We expand the source checkpoint into SST by copying each source layer into the corresponding layers of the Prefiller and the Decoder. The two towers share a single copy of the embedding, output normalization, and output head.

After expansion, both towers are trained jointly in BF16 using the source model's Muon~\cite{liu2025muon} and Adam~\cite{kingma2014adam} parameter groups. We retain the optimizer states for the Prefiller and the shared parameters and initialize fresh states for the Decoder. All groups continue on the source cosine learning-rate schedule, with the remaining decay extended over the continuation stage and no separate warmup for the Decoder. Appendix~\ref{app:schedule} gives the detailed settings.

\subsection{Cumulative training compute}
\label{sec:budget}

We compare the theoretical FLOPs of the effective architectures. For source and continuation token counts $D_s,D_c$ and forward FLOPs/token $f_s,f_c$, we use
\begin{equation}
 C_{\mathrm{train}}=3(D_s f_s+D_c f_c),
\label{eq:traincost}
\end{equation}
where the factor three approximates forward-plus-backward computation. The full cost of both training stages is included. The evaluated 47B classic checkpoint defines $\bref\approx5.16836\times10^{21}$ FLOPs. The final SST and 63B endpoints both use approximately 100\% of this budget. Appendix~\ref{app:coordinates} gives the theoretical accounting and training budgets.

\section{Results}
\label{sec:results}

\subsection{Training loss}
\label{sec:loss}

\begin{figure}[!htb]
\centering\includegraphics[width=\linewidth]{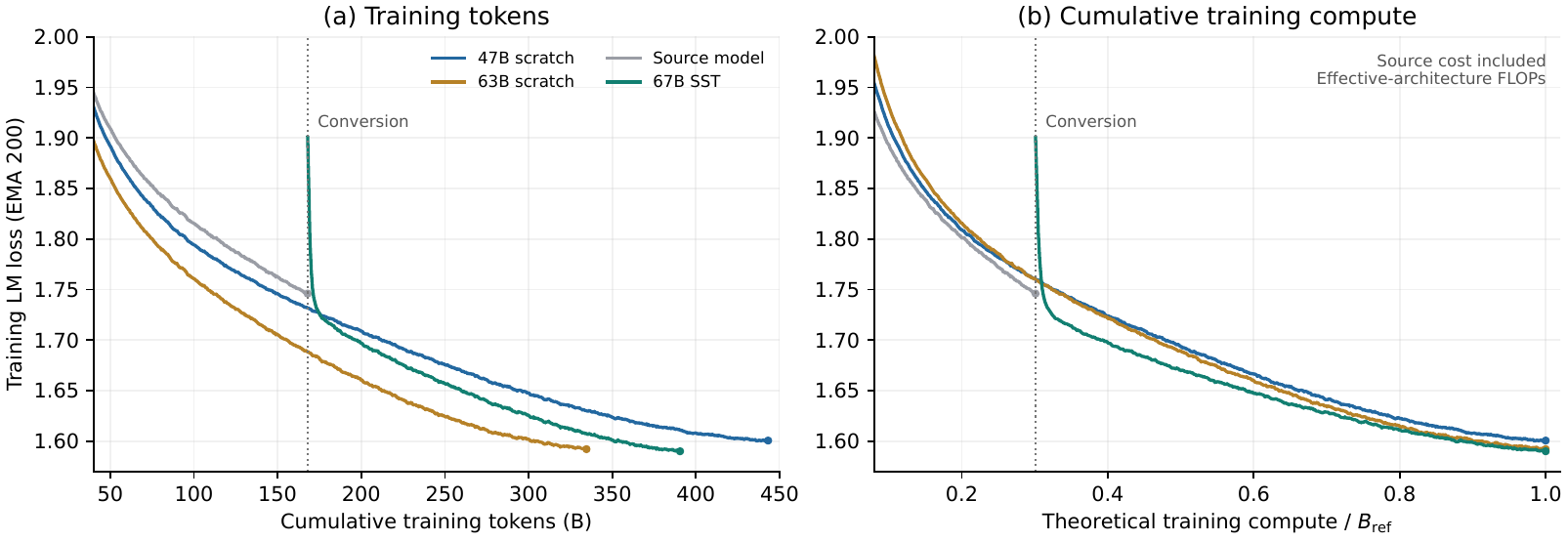}
\caption{Training LM loss (EMA-200) against tokens and theoretical cumulative compute. Conversion occurs at 167.98B tokens ($0.301\bref$).}
\label{fig:loss}
\end{figure}

SST's loss rises at conversion and then decreases through continuation (Figure~\ref{fig:loss}). At the final checkpoints, its training EMA-200 is 1.5900 after 390.54B tokens, versus 1.6006 for 47B classic after 443.16B tokens and 1.5921 for 63B classic after 334.62B tokens. SST therefore reaches lower training loss than both references at comparable cumulative training compute. Training budgets and loss reporting are given in Appendix~\ref{app:coordinates}.

\subsection{Task-level evaluation}

Table~\ref{tab:bmk} reports scores on seven downstream tasks (OpenBookQA~\cite{mihaylov2018openbookqa}, MMLU~\cite{hendrycks2020mmlu}, GSM8K~\cite{cobbe2021gsm8k}, MATH~\cite{hendrycks2021math}, HumanEval~\cite{chen2021humaneval}, MBPP~\cite{austin2021mbpp}, and BBH~\cite{suzgun2022bbh}) at the final checkpoints of all three models. These are the same checkpoints used for the training-loss and cumulative-compute comparisons. 
We also evaluate held-out token NLL on ARXIV using the same 1,024 frozen packed-4K samples for all three models.

Trained with an equal total of FLOPs, SST performs the best across the board. It shows that SST's lowest training loss indeed translates to the best downstream performance. SST also has the lowest estimated inference cost in the input-heavy regime analyzed in Section~\ref{sec:pd-cost}.


\begin{table}[!htb]
\centering\small
\caption{Scores (\%) on seven downstream tasks and held-out ARXIV token NLL (lower is better), evaluated at comparable cumulative theoretical training FLOPs, including both training stages for SST. OpenBookQA combines the validation and test splits; MMLU uses the test split, and MATH uses the full test set.}
\label{tab:bmk}
\begin{tabular}{@{}lrrr@{}}
\toprule
Task & 47B classic & 63B classic & 67B SST\\
\midrule
OpenBookQA & 68.50 & 71.00 & \textbf{75.00}\\
MMLU & 58.44 & 59.39 & \textbf{60.54}\\
GSM8K & 56.56 & 55.04 & \textbf{58.38}\\
MATH & 33.06 & 33.14 & \textbf{34.36}\\
HumanEval (few-shot) & 35.98 & 32.93 & \textbf{37.80}\\
MBPP (3-shot) & 50.40 & 50.80 & \textbf{51.40}\\
BBH & 50.36 & 52.56 & \textbf{52.74}\\
ARXIV NLL ($\downarrow$) & 1.4521 & 1.4424 & \textbf{1.4421}\\
\bottomrule
\end{tabular}

\end{table}




\subsection{Inference cost across prefill--decode mixes}
\label{sec:pd-cost}

Figure~\ref{fig:pd-cost} compares estimated inference cost across prefill:decode (P:D) cost mixes, using the proxy in Appendix~\ref{app:openrouter} and active-body counts in Appendix~\ref{app:params}. All models use the same P:D weights, with costs normalized to 47B classic. SST breaks even with 63B at 13.4:86.6 and with 47B at 65.5:34.5, becoming cheaper as the prefill share increases. At 75:25, its estimated cost is 6.7\% lower than 47B and 31.6\% lower than 63B.

\begin{figure}[!htb]
\centering\includegraphics[width=0.9\linewidth]{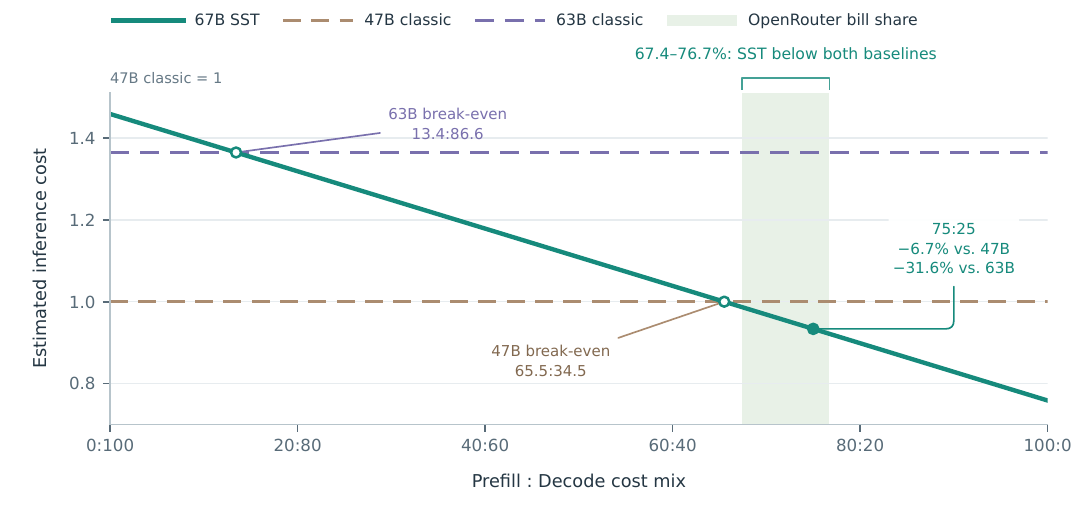}
\caption{Estimated inference cost across P:D cost mixes, normalized to 47B classic. Shading spans the six-model OpenRouter reference range; open circles mark break-even points and the filled circle marks 75:25.}
\label{fig:pd-cost}
\end{figure}

The six OpenRouter models in Table~\ref{tab:openrouter-intro} have uncached-input bill shares of 67.4--76.7\% (Appendix~\ref{app:openrouter}). Using these shares as reference prefill weights, SST is cheaper than both baselines throughout the highlighted range: 1.3--7.9\% below 47B and 27.7--32.5\% below 63B. These are analytical cost estimates, not measured serving speedups.
\section{Discussion}
\label{sec:connectivity-discussion}

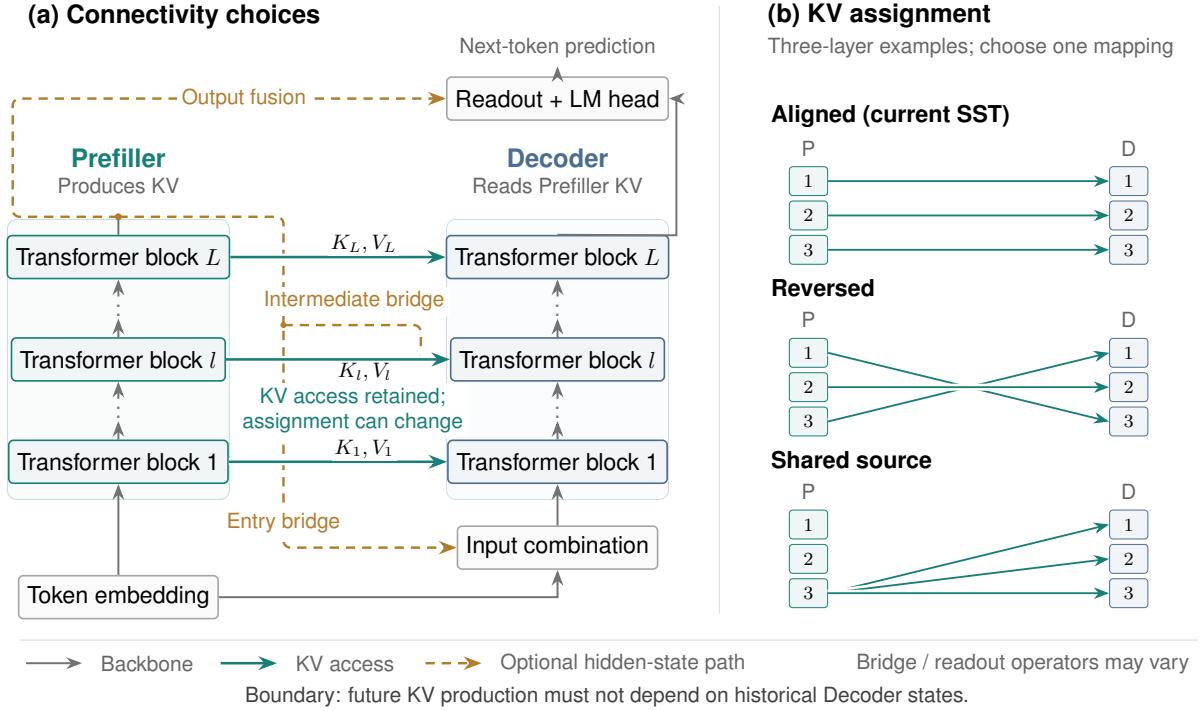
\begin{figure}[!htb]
\centering\resizebox{0.94\linewidth}{!}{
\definecolor{prefillercolor}{HTML}{177F77}
\definecolor{decodercolor}{HTML}{496E91}
\definecolor{optioncolor}{HTML}{B27C2D}
\begin{tikzpicture}[x=1cm,y=1cm,font=\sffamily\small,>=Stealth,
 box/.style={rounded corners=2pt,minimum width=2.8cm,minimum height=.65cm,align=center,line width=.7pt},
 pblock/.style={box,draw=prefillercolor,fill=prefillercolor!7},
 dblock/.style={box,draw=decodercolor,fill=decodercolor!7},
 neutral/.style={box,draw=black!35,fill=black!2},
 flow/.style={->,line width=.85pt,draw=black!55},
 kv/.style={->,line width=1.1pt,draw=prefillercolor,preaction={draw=white,line width=3pt}},
 optional/.style={->,rounded corners=3pt,dash pattern=on 4pt off 3pt,line width=1.05pt,draw=optioncolor},
 note/.style={font=\sffamily\footnotesize,text=black!60,align=center},
 tag/.style={font=\sffamily\footnotesize,text=optioncolor,fill=white,inner sep=2pt,align=center},
 psmall/.style={draw=prefillercolor,fill=prefillercolor!7,rounded corners=1.5pt,minimum width=.58cm,minimum height=.38cm,font=\sffamily\scriptsize},
 dsmall/.style={draw=decodercolor,fill=decodercolor!7,rounded corners=1.5pt,minimum width=.58cm,minimum height=.38cm,font=\sffamily\scriptsize}]

\path[use as bounding box] (-.05,-1.65) rectangle (18.15,9.1);
\node[anchor=west,font=\sffamily\bfseries] at (.1,8.8) {(a) Connectivity choices};
\node[anchor=west,font=\sffamily\bfseries] at (11.3,8.8) {(b) KV assignment};
\node[note,anchor=west] at (11.3,8.32) {Three-layer examples; choose one mapping};
\draw[black!15] (10.7,-.25)--(10.7,8.95);

\node[font=\sffamily\bfseries,text=prefillercolor] at (1.6,6.65) {Prefiller};
\node[font=\sffamily\bfseries,text=decodercolor] at (8.25,6.65) {Decoder};
\node[note] at (1.6,6.23) {Produces KV};
\node[note] at (8.25,6.23) {Reads Prefiller KV};
\draw[draw=prefillercolor!25,fill=prefillercolor!2,rounded corners=4pt] (-.08,1.48) rectangle (3.28,5.72);
\draw[draw=decodercolor!25,fill=decodercolor!2,rounded corners=4pt] (6.57,1.48) rectangle (9.93,5.72);
\node[pblock] (p1) at (1.6,2.05) {Transformer block 1};
\node[pblock] (pm) at (1.6,3.6) {Transformer block $l$};
\node[pblock] (pL) at (1.6,5.15) {Transformer block $L$};
\node[dblock] (d1) at (8.25,2.05) {Transformer block 1};
\node[dblock] (dm) at (8.25,3.6) {Transformer block $l$};
\node[dblock] (dL) at (8.25,5.15) {Transformer block $L$};
\foreach \x in {1.6,8.25}{
 \node[text=black!50] at (\x,2.83) {$\vdots$};
 \node[text=black!50] at (\x,4.38) {$\vdots$};
}

\node[neutral] (emb) at (1.6,.0) {Token embedding};
\node[neutral] (entry) at (8.25,.75) {Input combination};
\node[neutral,minimum width=3.15cm] (head) at (8.25,7.55) {Readout + LM head};
\node[note] (prediction) at (8.25,8.32) {Next-token prediction};
\draw[flow] (emb)--(p1);
\draw[flow] (emb.east)--(8.25,0)--(entry.south);
\draw[flow] (entry)--(d1);
\foreach \b in {p1,pm,d1,dm}{\draw[flow] (\b.north)--++(0,.28);}
\foreach \b in {pm,pL,dm,dL}{\draw[flow] (\b.south)++(0,-.26)--(\b.south);}
\draw[flow] (dL.north)--(10.05,5.48)--(10.05,7.55)--(head.east);
\draw[flow] (head)--(prediction);

\coordinate (pf) at (1.6,5.76);
\draw[flow,-] (pL.north)--(pf);
\fill[optioncolor] (pf) circle (1.5pt);
\draw[optional] (pf)--(4.1,5.76)--(4.1,.75)--(entry.west);
\node[tag] at (4.1,1.12) {Entry bridge};
\draw[optional] (4.1,4.12)--(6.18,4.12)--(6.18,3.6)--(dm.west);
\fill[optioncolor] (4.1,4.12) circle (1.3pt);
\node[tag] at (5.17,4.48) {Intermediate bridge};
\draw[optional] (pf)--(-.0,5.76)--(-.0,7.55)--(head.west);
\node[tag] at (3.5,7.55) {Output fusion};

\draw[kv] (p1.east)--(d1.west) node[pos=.62,above,fill=white,inner sep=1pt,font=\footnotesize] {$K_1,V_1$};
\draw[kv] (pm.east)--(dm.west) node[pos=.62,below,fill=white,inner sep=1pt,font=\footnotesize] {$K_l,V_l$};
\draw[kv] (pL.east)--(dL.west) node[pos=.62,above,fill=white,inner sep=1pt,font=\footnotesize] {$K_L,V_L$};
\node[note,text=prefillercolor,fill=white,inner sep=2pt] at (5.15,2.83) {KV access retained;\\assignment can change};

\foreach \name/\yy/\title in {a/6.45/Aligned (current SST),r/3.85/Reversed,s/1.25/Shared source}{
 \node[font=\sffamily\small\bfseries,anchor=west] at (11.35,\yy+.83) {\title};
 \node[note] at (12.05,\yy+.35) {P};
 \node[note] at (16.9,\yy+.35) {D};
 \foreach \idx/\dy/\lab in {1/0/1,2/-.52/2,3/-1.04/3}{
  \node[psmall] (\name p\idx) at (12.05,\yy+\dy-.15) {$\lab$};
  \node[dsmall] (\name d\idx) at (16.9,\yy+\dy-.15) {$\lab$};
 }
}
\foreach \i in {1,2,3}{\draw[kv,line width=.8pt] (ap\i.east)--(ad\i.west);}
\draw[kv,line width=.8pt] (rp1.east)--(rd3.west);
\draw[kv,line width=.8pt] (rp3.east)--(rd1.west);
\draw[kv,line width=.8pt] (rp2.east)--(rd2.west);
\foreach \i in {1,2,3}{\draw[kv,line width=.8pt] (sp3.east)--(sd\i.west);}

\draw[black!15] (.1,-.65)--(17.95,-.65);
\draw[flow] (.2,-1.0)--(1.05,-1.0);
\node[note,anchor=west] at (1.2,-1.0) {Backbone};
\draw[kv] (3.15,-1.0)--(4,-1.0);
\node[note,anchor=west] at (4.15,-1.0) {KV access};
\draw[optional] (6.25,-1.0)--(7.1,-1.0);
\node[note,anchor=west] at (7.25,-1.0) {Optional hidden-state path};
\node[note,anchor=east] at (17.95,-1.0) {Bridge / readout operators may vary};
\node[font=\sffamily\footnotesize,text=black!75] at (9,-1.5)
 {Boundary: future KV production must not depend on historical Decoder states.};
\end{tikzpicture}}
\caption{Connectivity choices within KITE. (a) An SST-style backbone with optional hidden-state connections at the Decoder entry, intermediate layers, and output readout. Dashed paths indicate design choices, not connections that must be enabled together. (b) Three illustrative KV assignments: aligned layer-wise reuse, reversed pairing, and reuse of one Prefiller KV source across multiple Decoder layers. The KV-producing computation remains on the Prefiller side; the illustrated choices are not assumed to have equal cost or quality.}
\label{fig:design-space}
\end{figure}

SST provides one concrete realization of KITE, rather than a unique architectural prescription. Its aligned layer-wise KV reuse, entry bridge, and Decoder-only readout are specific design choices. The central requirement is that the added capacity does not extend the KV-producing computation or introduce dependencies that require historical Decoder states. Figure~\ref{fig:design-space} illustrates several connectivity choices within this framework.

The assignment between Prefiller KV and Decoder layers need not be one-to-one or order-preserving. For example, Decoder layers could read Prefiller KV in reversed layer order, or multiple Decoder layers could reuse a single KV source. Hidden-state connections offer another degree of freedom: Prefiller representations may enter the Decoder at its input or at intermediate layers, and may also contribute to the final readout. The bridge operators, as well as the depth and width of the two components, can likewise vary. These choices must preserve causal access and compatible interfaces, but need not have the same computational cost or model quality.

Our experiments explore only a limited part of this design space. The results reported in this paper establish the performance of the selected SST configuration, rather than identify an optimal connectivity pattern. Further exploration can examine how KV assignment, hidden-state connections, and capacity allocation affect the balance between model quality, training computation, and inference cost. KITE provides a common setting for studying these choices without fixing the architecture to the particular connections used in SST.

\FloatBarrier
\section{Related work}
\label{sec:discussion}

Net2Net studies function-preserving model growth~\cite{chen2015net2net}; SPARKLING examines signal preservation and symmetry breaking in width-progressive learning~\cite{yu2026sparkling}. Dense-to-MoE upcycling reuses dense checkpoints to initialize sparse expert models~\cite{komatsuzaki2022upcycling,he2024upcycling}. SST instead adds the Decoder to an already sparse source partway through training. Its aligned initialization changes the prediction graph and is followed by joint adaptation.

Multi-query attention~\cite{shazeer2019mqa} and grouped-query attention~\cite{ainslie2023gqa} reduce KV storage by sharing KV heads across query heads. Cross-layer attention extends sharing across adjacent layers~\cite{brandon2024cla}. These approaches reduce the stored attention state; KITE targets capacity expansion without extending the KV-producing path.

YOCO separates cache-producing and cache-consuming computation and exploits this separation for prefill early exit~\cite{sun2024yoco}. SST applies this execution pattern to staged model growth. YOCO's cross-decoder shares a global KV representation; SST reads aligned layer-wise KV and is constructed through source-to-target expansion.

Mixture-of-Recursions explores recursive KV sharing, reusing the first recursion's KV across subsequent recursions~\cite{bae2025mixtureofrecursions}. From an architectural perspective, such fixed-KV looping can be viewed as a parameter-tied special case of the family underlying KITE. SST instead uses independent Prefiller and Decoder parameters, expanding capacity during training without extending the KV-producing path.

\section{Conclusion}

KITE expands a model during training by adding capacity that reuses KV, leaving bulk-prefill computation source-sized. SST provides a simple instance of this route. With both training stages counted, it reaches lower training loss than the 47B and 63B classic baselines at comparable theoretical training compute. It also scores higher on several downstream tasks. Relative to the 47B baseline, SST trades lower bulk-prefill computation for higher decode computation; its inference-cost proxy is lower in prefill-heavy workloads. The larger 63B reference has a higher inference proxy and slightly higher training loss. These observations support further investigation of the combined scaling route, without establishing a scaling law, a causal benefit from each ingredient, or measured serving speedup.

\FloatBarrier
\appendix
\section{OpenRouter workload statistics and inference proxy}
\label{app:openrouter}

\paragraph{Data and coverage.}
Table~\ref{tab:openrouter-intro} summarizes six selected models using public OpenRouter daily activity retrieved on September 21, 2026. We aggregate available standard and Batch activity within August 22--September 20 UTC, without extrapolating missing days. Both Astra variants have 17 observed days (September 4--20), and Gemini 3.8 Flash has 19 (September 2--20). The other models have 30 standard-activity days; Claude Fable 5 has only 10 Batch days (August 22--31). This selected traffic is not separately labeled as agentic usage.

\paragraph{Token and charge accounting.}
Let $P$ be total input tokens, $C$ cache-read tokens, $U=P-C$ uncached input tokens, and $D$ output tokens, including reasoning. The two token ratios are $P/D$ and $U/D$. For each variant $v$, let $a_v$ and $o_v$ be the September 21 base input and output prices per million tokens. We estimate
\begin{equation}
 A=10^{-6}\sum_v U_v a_v,\qquad
 O=10^{-6}\sum_v D_v o_v,\qquad
 s_{\mathrm{uncached}}=\frac{A}{A+O}.
\end{equation}
Standard and Batch charges are calculated separately before summation. Cache-read charges are excluded from both numerator and denominator. These are base-price estimates, not invoices; cache-write premiums, context-length and provider-specific pricing, historical price changes, and non-token fees are not modeled.

\begin{table}[!htb]
\centering\small
\setlength{\tabcolsep}{5pt}
\caption{Components underlying Table~\ref{tab:openrouter-intro}. Token counts are in billions; estimated charges are in millions of USD. $U$ is uncached input, $C$ cached input, and $D$ output. Ratios use unrounded totals.}
\label{tab:openrouter-traffic}
\begin{tabular}{@{}lrrrrr@{}}
\toprule
Model & $U$ (B) & $C$ (B) & $D$ (B) & $A$ (\$M) & $O$ (\$M)\\
\midrule
GPT-6 Astra & 356.199 & 2232.435 & 24.323 & 3.5600 & 1.2135\\
GPT-6 Astra Pro & 45.611 & 159.946 & 3.255 & 0.4554 & 0.1613\\
Claude Fable 5 & 159.115 & 460.103 & 9.770 & 1.5895 & 0.4822\\
Claude Sonnet 5 & 1126.045 & 4434.707 & 92.872 & 2.2514 & 0.9279\\
Gemini 3.7 Flash & 2100.037 & 5722.111 & 207.001 & 1.5666 & 0.7570\\
Gemini 3.8 Flash & 1299.288 & 4405.218 & 123.644 & 0.9714 & 0.4598\\
\bottomrule
\end{tabular}
\smallnote{Standard input/output rates (USD per million tokens) are 10/50 for Astra, Astra Pro and Fable 5; 2/10 for Sonnet 5; and 0.75/3.75 for both Gemini models. Batch rates are half the corresponding standard rates.}
\end{table}

\paragraph{Sources.}
Activity comes from OpenRouter's public model-activity API: \href{https://openrouter.ai/api/frontend/v1/stats/model-activity?permaslug=openai\%2Fgpt-6-astra-20260903&variant=standard}{Astra}, \href{https://openrouter.ai/api/frontend/v1/stats/model-activity?permaslug=openai\%2Fgpt-6-astra-pro-20260903&variant=standard}{Astra Pro}, \href{https://openrouter.ai/api/frontend/v1/stats/model-activity?permaslug=anthropic\%2Fclaude-5-fable-20260609&variant=standard}{Fable 5}, \href{https://openrouter.ai/api/frontend/v1/stats/model-activity?permaslug=anthropic\%2Fclaude-sonnet-5-20260630&variant=standard}{Sonnet 5}, \href{https://openrouter.ai/api/frontend/v1/stats/model-activity?permaslug=google\%2Fgemini-3.7-flash-20260813&variant=standard}{Gemini 3.7 Flash}, and \href{https://openrouter.ai/api/frontend/v1/stats/model-activity?permaslug=google\%2Fgemini-3.8-flash-20260902&variant=standard}{Gemini 3.8 Flash}; Batch uses the same model identifier with \code{variant=batch}. Prices use the \href{https://openrouter.ai/api/v1/models}{OpenRouter catalog} snapshot; token definitions follow its \href{https://openrouter.ai/docs/cookbook/administration/usage-accounting}{usage-accounting} and \href{https://openrouter.ai/docs/guides/best-practices/prompt-caching}{prompt-caching} documentation. All snapshots were retrieved on September 21, 2026.

\paragraph{Relation to the inference proxy.}
These statistics motivate input-heavy inference but do not measure GPU costs. We apply the same prefill:decode cost weights to all models. For prefill weight $w$, the proxy is $wp+(1-w)d$, where $p$ and $d$ are bulk-prefill and decode active-body parameter counts normalized to 47B classic. Thus $w=0.75$ means a 75:25 P:D cost mix. Figure~\ref{fig:scaling-route} uses this mix, giving 0.933 for SST and 1.365 for 63B classic, compared with 1 for 47B classic: SST is 6.7\% below 47B and 31.6\% below 63B. Figure~\ref{fig:pd-cost} varies the same cost weights. These are analytical proxy comparisons, not measured serving speedups.

\section{Parameter accounting}
\label{app:params}

Parameter counts follow the configured tensor shapes, checked against the exported weight inventory. SST has two sets of Transformer blocks and one shared embedding, output norm, and output head. Only the Prefiller contains K/V projections and K normalization in the effective architecture. The entry and final-state RMS operations have no trainable parameters.

\begin{table}[!htb]
\centering\small
\caption{Detailed geometry. Source/SST dimensions apply to each tower; both have two initial dense FFN layers. RoPE denotes rotary position embeddings~\cite{su2021roformer}.}
\begin{tabular}{@{}lrrr@{}}
\toprule
Setting & Source / SST & 47B classic & 63B classic\\
\midrule
Dense FFN width & 4608 & 5120 & 5632\\
Routed / shared expert width & 576 / 576 & 640 / 640 & 704 / 704\\
Head dimension & 128 & 128 & 128\\
Full / sliding RoPE dimension & 64 / 128 & 64 / 128 & 64 / 128\\
Full / sliding Q heads & 24 / 40 & 24 / 40 & 32 / 48\\
KV heads & 8 & 8 & 8\\
\bottomrule
\end{tabular}
\end{table}

Total parameters count the effective architecture. Active body parameters exclude embedding/head and count Top-8 routed experts per MoE layer, shared experts, routers, and the remaining body weights needed for one token. These are per-token parameter counts, not measured execution costs.

\begin{table}[!htb]
\centering\small
\caption{Architectural parameter counts in billions, rounded to three decimal places.}
\begin{tabular}{@{}lrrr@{}}
\toprule
Quantity & SST & 47B classic & 63B classic\\
\midrule
Total, including embedding/head & 66.959 & 46.727 & 62.691\\
Embedding + output head & 0.594 & 0.660 & 0.725\\
Total body & 66.365 & 46.068 & 61.966\\
Decode active body & 2.155 & 1.477 & 2.016\\
Prefiller-only active body & 1.120 & -- & --\\
\bottomrule
\end{tabular}
\end{table}

The exported vocabulary has 128,815 rows, without training padding. Excluding embedding and head from the active-body counts avoids treating the full embedding matrix as a per-token multiplication. The output head still contributes prompt-boundary and decode work and must be included in measured serving cost.

\section{Training budgets and loss reporting}
\label{app:coordinates}

\begin{table}[!htb]
\centering\small
\caption{Training budgets at the final checkpoints. Theoretical training FLOPs are normalized to the 47B classic reference.}
\begin{tabular}{@{}lrr@{}}
\toprule
Model & Training tokens (B) & Relative training FLOPs\\
\midrule
47B classic & 443.16 & 1.000\\
63B classic & 334.62 & $\approx 1.000$\\
67B SST & 390.54 & $\approx 1.000$\\
\bottomrule
\end{tabular}
\end{table}

SST's total includes 167.98B source-stage tokens and 222.55B continuation tokens; both stages are included in its training compute.

The loss curves end at the evaluated final checkpoints. Training-loss curves use an exponential moving average with span 200 (\code{adjust=False}), computed from the recorded observations without interpolation. The EMA is restarted after SST conversion and computed before plot subsampling, with the final observation retained.

The evaluated 47B classic checkpoint defines $\bref\approx5.16836\times10^{21}$ FLOPs. Forward costs are approximately 3.887488B FLOPs/token for 47B classic, 5.148513B for 63B classic, 3.087252B for the source, and 5.410678B for SST. SST retains both towers' attention and FFN/MoE computation but only one embedding, output head, and set of KV projections.

All forward coefficients use sequence length 4096, sliding-window size 512, and the padded training vocabulary of 128,896, with multiply-add counted as two FLOPs. They include attention projections and matrix products, embedding, output head, FFN/MoE, routers, and the estimator's activation terms. Normalization, RoPE, residual and bridge operations, rematerialization, optimizer, loss/auxiliary loss, conversion, communication, and checkpoint I/O are excluded. The same theoretical accounting convention is used for all models.

\section{Training recipe and continuation scheduler}
\label{app:schedule}

Continued training uses native BF16 parameters with Muon and Adam groups. Muon uses six Newton--Schulz steps, with packed attention split by head and GLU split by projection type; Adam uses $\epsilon=10^{-15}$. The source peak base learning rate is approximately $1.03302\times10^{-3}$. The continuation scheduler uses a base weight decay of 0.15 with the inherited parameter-group multipliers. The source/SST/63B token batch is 12,582,912; the 47B token batch is 14,680,064.

SST continuation uses 128 NVIDIA H800 GPUs with expert parallelism of size 8, a sequence length of 4,096, and a global batch size of 3,072 sequences. It runs for 17,687 updates after 13,350 source-training updates, giving 31,037 updates in total.

The parameter-free RMS operations at the Decoder input and final state accumulate in FP32 and return the input dtype; the original learned output norm is applied after the final-state RMS operation.

Let $u$ be the number of continuation tokens consumed, $U=222{,}553{,}964{,}544$ the configured continuation horizon, and $q_0$ the source's normalized cosine phase at conversion. SST uses
\begin{align}
 q(u)&=q_0+(1-q_0)\min(u/U,1),\\
 \eta(u)&=\eta_{\min}+(\eta_{\mathrm{anchor}}-\eta_{\min})
 \frac{1+\cos(\pi q(u))}{1+\cos(\pi q_0)}.
\end{align}
The source schedule warms up over 12.582912B tokens and decays until 335.9605248B tokens. Conversion occurs at 167.9818752B tokens. Thus $q_0$ is measured within the post-warmup decay interval, not by dividing the conversion position by the whole horizon. The learning-rate constants are
\begin{align*}
 \eta_{\mathrm{anchor}}&\approx5.96548\times10^{-4},\\
 \eta_{\min}&\approx1.03302\times10^{-4}.
\end{align*}
Parameter-group multipliers remain applicable to the common base learning rate.

\FloatBarrier
\begingroup
\small
\setlength{\bibsep}{2pt}
\bibliographystyle{plainnat}
\bibliography{references}
\endgroup
\end{document}